%% file: manuscript.tex
\pdfoutput=1

\documentclass[11pt]{article}

\usepackage[final]{acl}

\usepackage{times}
\usepackage{latexsym}

\usepackage{graphicx}
\usepackage{longtable}
\usepackage{booktabs}
\usepackage{subcaption}
\usepackage{tabularx}
\usepackage{supertabular}
\usepackage{float}
\usepackage{color,soul}

\usepackage[T1]{fontenc}

\usepackage[utf8]{inputenc}

\usepackage{microtype}

\usepackage{inconsolata}

\title{Transformer verbatim in-context retrieval across time and scale}

\author{Kristijan Armeni \\
  Johns Hopkins University \\
  \texttt{karmeni1@jhu.edu} \\\And
  Marko Pranjić \\
  Jozef Stefan Institute \\
  Jozef Stefan International\\Postgraduate School\\
  \texttt{marko.pranjic@ijs.si} \\\And
  Senja Pollak \\
  Jozef Stefan Institute \\
  \texttt{senja.pollak@ijs.si} \\}

\begin{document}
\maketitle
\begin{abstract}

To predict upcoming text, language models must in some cases retrieve in-context information verbatim.  In this report, we investigated how the ability of language models to retrieve arbitrary in-context nouns developed during training (across time) and as language models trained on the same dataset increase in size (across scale). We then asked whether learning of in-context retrieval correlates with learning of more challenging zero-shot benchmarks. Furthermore, inspired by semantic effects in human short-term memory, we evaluated the retrieval with respect to a major semantic component of target nouns, namely whether they denote a concrete or abstract entity, as rated by humans. We show that verbatim in-context retrieval developed in a sudden transition early in the training process, after about 1\% of the training tokens. This was observed across model sizes (from 14M and up to 12B parameters), and the transition occurred slightly later for the two smallest models. We further found that the development of verbatim in-context retrieval is positively correlated with the learning of zero-shot benchmarks. Around the transition point, all models showed the advantage of retrieving concrete nouns as opposed to abstract nouns. In all but two smallest models, the advantage dissipated away toward the end of training.

\end{abstract}

\section{Introduction}

In language models (LMs), successful prediction of upcoming words depends on in-context information. For example, when given the context prompt ``\textit{The novel's plot and symbolism are centered around three objects: a centipede, a parachute, and a waterfall. The first and most important object in the list is the \_\_\_}'', an LM must retrieve the noun (\textit{centipede}) out of all in-context tokens to correctly predict the continuation. In human cognitive science, this ability to flexibly retrieve items from recent context is known as short-term memory and is believed to be the core computation underlying human cognition \citep{baddeley_working_2003a}.

\begin{figure}

\includegraphics[width=1\linewidth]{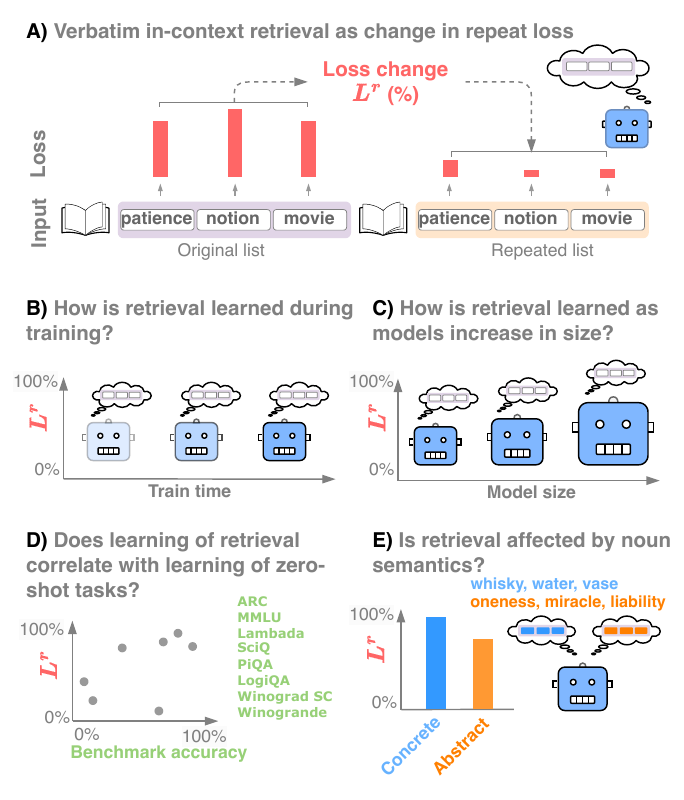}
\caption{Overview of the approach and experiments.}
\label{fig:overview}
\vspace{-10pt}
\end{figure}

Recently, \citet{armeni_characterizing_2022} showed that a transformer language model \citep[GPT-2,][]{brown_language_2020} develops such flexible short-term memory --- it was able to retrieve the identity and ordering of lists of \textit{arbitrary} nouns from recent context (Fig. \ref{fig:overview}, A), even though retrieval of arbitrary in-context information is not the explicit objective of LMs (as opposed to dedicated models of short-term memory,  e.g. \citealt{oberauer_benchmarks_2018}). Yet, studying retrieval in a single \textit{fully-trained} model on \textit{arbitrary} nouns neglects three further dimensions of the capacity: how it is learned, how learning of this dedicated capacity relates to models' learning of other tasks, and the semantics of retrieved nouns.

First, studying \textit{learning trajectories} of LM capacities offers complementary insights to studying only performance of fully-trained models \citep[e.g.][]{chen_sudden_2024}. Previous work on LM learning trajectories showed that transformers learn next-token prediction by undergoing a \textit{sudden transition} (``phase change'') early during training, which coincides with the development of attention heads that attend to repeated tokens \citep{olsson_-context_2022}. Does verbatim retrieval follow a similar learning trajectory?

Second, the ability to retrieve and predict in-context tokens verbatim (i.e. identity-based matching) can be viewed as a rudimentary form of the more flexible zero-shot learning, where the relevant in-context information is not necessarily given verbatim and must possibly be retrieved based on fuzzy, similarity-based matching \cite{olsson_-context_2022}. How does successful learning of verbatim retrieval relate to LM's zero-shot performance on more challenging benchmark tasks?

Third, while the successful retrieval of arbitrary nouns underscores the flexibility of transformer short-term memory, this approach neglects that the lexicon of natural language is not a set of unorganized, arbitrary words --- instead, it has semantic structure. Two prominent semantic categories are \textit{concrete} and \textit{abstract }nouns. Concrete nouns (e.g. ``hammer'') have sensory referents, whereas abstract nouns (e.g. ``justice'') do not have a straightforward sensory component. Word concreteness affects human cognitive processing. Children typically acquire concrete words, especially nouns, earlier than abstract words \citep{gleitman_hard_2005}. In certain short-term memory paradigms, humans are better at recalling concrete than abstract words \citep{taylor_mechanisms_2019}. Importantly, the two word categories differ also in their distributional properties: concrete words occur in a semantically narrower range of contexts compared to abstract words \citep{schulte_im_walde_distributional_2022}. Is the transformer retrieval affected by whether nouns refer to concrete vs. abstract entities?

To address these questions, we evaluated verbatim in-context retrieval on the Pythia suite of language models \citep{biderman_pythia_2023}. Leveraging the fact that the suite includes pretrained LMs ranging from 14M to 12B parameters in scale and their intermediate training checkpoints across the entire learning epoch, we evaluated how retrieval develops over the course of training and across model sizes (Fig. \ref{fig:overview}, B and C). Additionally, the Pythia suite contains zero-shot evaluations on various benchmarks for each LM checkpoint. To test how in-context retrieval relates to LM's zero-shot performance, we correlated the learning trajectory of the retrieval against the learning trajectories on zero-shot benchmarks (Fig. \ref{fig:overview}, D). Finally, to test the role of noun semantics for in-context retrieval, we evaluated how noun concreteness, as rated by human participants \cite{brysbaert_concreteness_2014}, affected retrieval over the course of training (Fig. \ref{fig:overview}, E).

The main contributions of the current work are: \textbf{a)} In all models, verbatim retrieval developed in a sudden transition early during training, after about 1\% training tokens elapsed, and remained constant during the rest of training, \textbf{b)} learning of verbatim retrieval was positively correlated with learning of zero-shot task performance, and \textbf{c)} around the transition point, LMs showed an advantage to retrieve concrete rather than abstract nouns. This advantage almost entirely diminished towards the end of training.

\section{Related work}

Several recent studies investigated the behavior of LMs in domain of either verbatim or in-context retrieval more generally. \citet{armeni_characterizing_2022} developed a paradigm to test the short-term memory ability (in-context retrieval) of LMs. They showed that GPT-2 can retrieve the identity and ordering of repeated arbitrary nouns, but have only tested a single fully-trained LM and did not investigate learning trajectories. \citet{vaidya_humans_2023} compared LM (GPT-2) and human word prediction performance on spans of repeated text. They reported that LMs' next word prediction performance diverges from human performance on subsequent repetitions. They showed that GPT-2 performance aligned better with humans if its attention heads had a bias towards recent context. \citet{yu_characterizing_2023} investigated in-context retrieval of facts (e.g. retrieval of the capital city given a country name) and how such retrieval was affected by the pretraining statistics of retrieved facts. They showed that LMs (Pythia) could override retrieval of (counterfactual) in-context information and instead retrieved the fact that has a higher frequency of occurrence in training data (e.g., even when given the in-context counterfactual ``\textit{The capital city of Poland is London}'' they tend to predict the statistically more likely ``\textit{Warszaw}'').

The current report is also related to the recent work on LM interpretability and the role of attention heads in specific forms of retrieval. Several studies \citep{elhage_mathematical_2021, olsson_-context_2022, wang_interpretability_2023, yu_characterizing_2023} have identified circuits of attention heads that detect repeated in-context tokens and their previous continuations; the computations governing the behavior investigated presently. These studies focused either on how such attention mechanisms are learned and how they affect generic next-word prediction \citep{elhage_mathematical_2021, olsson_-context_2022} or how these attention mechanisms govern the retrieval of proper nouns as direct objects in sentences \citep{wang_interpretability_2022} or factual knowledge \citep[direct objects,][]{ yu_characterizing_2023}.

Here, we complement these lines of work and investigate retrieval as the ability of LMs to retrieve lists of arbitrary combinations of common nouns (unlikely seen co-occuring during training) and their semantic properties.

\begin{table*}[!ht]
    \centering
    \footnotesize
    \begin{tabular}{llll}
    \toprule
         Task&  Domain& Reference\\
         \midrule
         AI2 Reasoning Challenge (ARC) & Multiple choice science exams & \citet{clark_think_2018} \\
         Lambada& Discourse-based word prediction & \citet{paperno_lambada_2016} \\
         LogiQA& Logical reasoning & \citet{liu_logiqa_2020} \\
         Massive multitask lang. understanding (MMLU)& Exam knowledge across diverse domains & \citet{hendrycks_measuring_2021} \\
         PiQA & Physical common-sense reasoning & \citet{bisk_piqa_2020} \\
         SciQ& Scientific knowledge & \citet{welbl_crowdsourcing_2017} \\
         Winograd schema challenge (WSC) & Common-sense reasoning &\citet{levesque_winograd_2012} \\
         Winogrande & Common-sense reasoning &\citet{sakaguchi_winogrande_2021} \\
    \bottomrule
    \end{tabular}
    \caption{The benchmark tasks used to compute in learning trajectory correlations in Fig. \ref{fig:benchmark_corr}.}
    \label{tab:benchmarks}
\end{table*}

\section{Methods} \label{sec:exp_scale}

\subsection{Verbatim retrieval paradigm}

We used the verbatim retrieval paradigm introduced by \citet{armeni_characterizing_2022}. Here, LMs process a short vignette in English where a list of three arbitrary nouns is repeated twice:

\vspace{10pt}

\noindent\fbox{%
    \parbox{0.95\linewidth}{%
        \textit{Mary read a list of words: \textbf{patience, notion, movie}. After the meeting, she took a break and had a cup of coffee. When she got back, she read the list again: \textbf{patience, notion, movie}.}
    }%
}

\vspace{10pt}

We refer to the first list of nouns as original list and the second one as repeated list. This setup allows us to test how the LM behavior (as reflected in LM loss, see below) changes as the LM encounters the repeated list. The paradigm (retrieval of arbitrary lists of words) is broadly inspired by benchmarks for testing models of human working memory \citep[][]{oberauer_benchmarks_2018}. Whereas human participants can be tested by just being presented with lists of nouns alone, our paradigm is formatted such that it is more suited to be used as input to LMs: contextualized in a simple, but plausible natural language vignette.

\subsection{Quantifying verbatim retrieval} 

\paragraph{Change in repeat loss ($L^r$)} Following \citet{armeni_characterizing_2022}, we operationalized retrieval as a change in LM loss on repeated nouns. Specifically, we computed the ratio in LM $\mathrm{loss} = -\log_2{P(w_t|w_{1}, ..., w_{t-1}}$) between each noun in the original list and its repetition $k$ tokens later: $\mathrm{loss\ ratio}^{noun} = \frac{\mathrm{loss}(\mathrm{noun}_{i+k})}{\mathrm{loss}(\mathrm{noun}_i)}$. The loss ratio per list was obtained by averaging the noun-specific loss ratios over the three nouns in a list. A loss ratio $< 1$ indicates that the loss to the \textit{same tokens} has decreased (that is, the LM expected the token to repeat) and is taken as evidence of verbatim retrieval.

To quantify retrieval as increasing with better performance, we report it as repeat loss change $L^r = 1 - \mathrm{loss\ ratio}$, expressed as percentage. In this way, a 0\% change in repeat loss indicates no retrieval whereas a change towards 100\% indicates evidence towards (perfect) retrieval. Importantly, repeat loss change is a continuous measure of in-context retrieval, baselined against the LM loss at the beginning of the sequence which facilitates comparison across models (e.g. models that show different baseline loss as expected over the course of training and across scale) and across different types of inputs.

\subsection{Language models}

\paragraph{Pythia suite} To evaluate retrieval over the course of training and across scale (see Section \ref{sec:experiments} below), we used the publicly-available pretrained LM checkpoints released as part of the Pythia language modeling suite \cite{biderman_pythia_2023}.\footnote{\url{https://github.com/EleutherAI/pythia}} Pythia is a suite of decoder-only autoregressive transformer LMs spanning from 14M to 12B parameters in size together with 144 intermediate checkpoints stored during training. The models were trained on the Pile dataset \cite{gao_pile_2020}, an English-only corpus for training large-scale LMs containing texts from 22 sources \cite[for example, Common Crawl, Wikipedia, Project Gutenberg, Books3, arXiv etc., see][for details]{biderman_datasheet_2022}. The model checkpoints used in this report were trained on the version of the dataset containing approximately 300B tokens. For the full architecture and training details, readers are referred to the original report \cite{biderman_pythia_2023}.

In our experiments, we evaluated the following model sizes: \{14M, 31M, 70M, 160M, 410M, 1B, 6.9B, 12B\} at 18 training checkpoints spanning 6 orders of magnitude across the training steps (in number of training tokens, $10^6,...,10^{11}$) from the initialized to the final fully-trained model\footnote{Specifically we evaluated the checkpoints from the following training steps: \{0, 1, 4, 32, 128, 256, 512, 1000, 2000, 3000, 4000, 8000, 10000, 30000, 40000, 50000, 100000, 143000\}. A single step contained 2,097,152 tokens \cite{biderman_pythia_2023}.}. All model checkpoints were accessed through the HuggingFace Transformers library \cite{wolf_transformers_2020}. 

\subsection{Experiments} \label{sec:experiments}

\paragraph{Experiment 1: Retrieval of arbitrary nouns across time and scale} In the first experiment, word lists in the vignette were constructed by randomly sampling nouns from the Toronto word pool\footnote{\url{http://memory.psych.upenn.edu/files/wordpools/nouns.txt}} as used in \citet{armeni_characterizing_2022}. Noun lists in the set (23 lists of 10 nouns) were constructed such that each noun was tested in all 10 possible ordinal positions in the list (e.g. ``patience, notion, movie'', ``notion, movie, patience'', etc.) to control for any position-specific retrieval effects. This procedure resulted in the final stimulus set that contained $N = 230$ samples of vignettes. In the present experiment, we used the version of the stimulus set where the list length was capped at 3 nouns.

\begin{figure*}[!ht]
\centering
    \begin{subfigure}[t]{0.55\textwidth}
        \includegraphics[width=\textwidth]{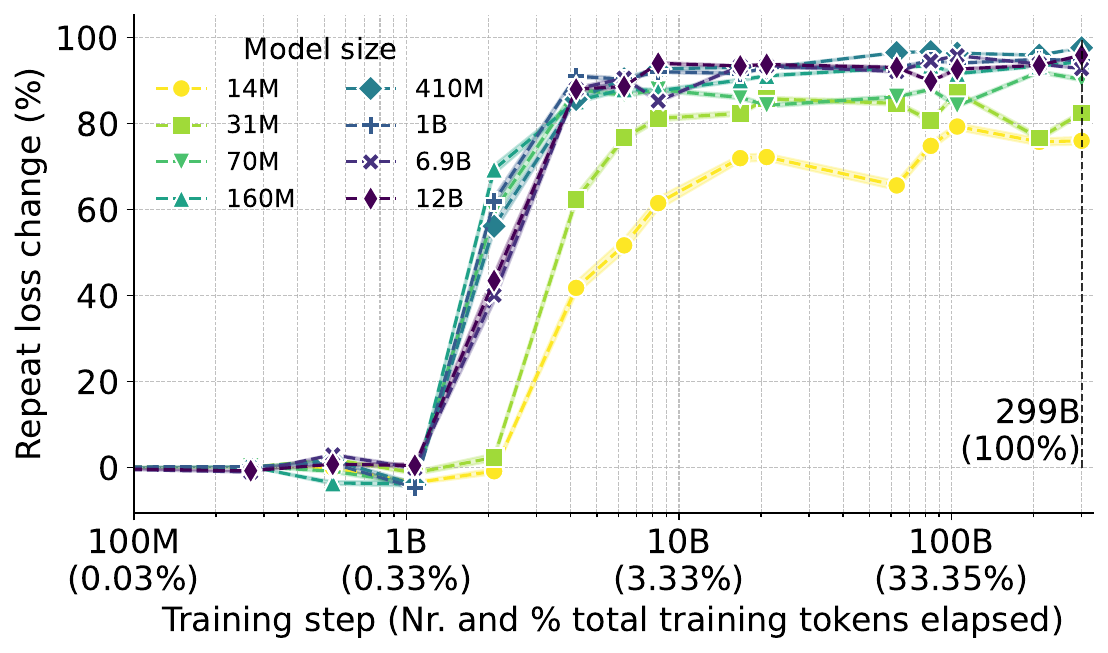}
        \caption{Capacity to retrieve verbatim repetitions of arbitrary nouns (average across three nouns in a list) is learned early during LM training and predominantly conserved across scale.}
        \label{fig:memory:dynamics}
    \end{subfigure}
    \hspace{30pt}
    \begin{subfigure}[t]{0.27\textwidth}
        \includegraphics[width=\textwidth]{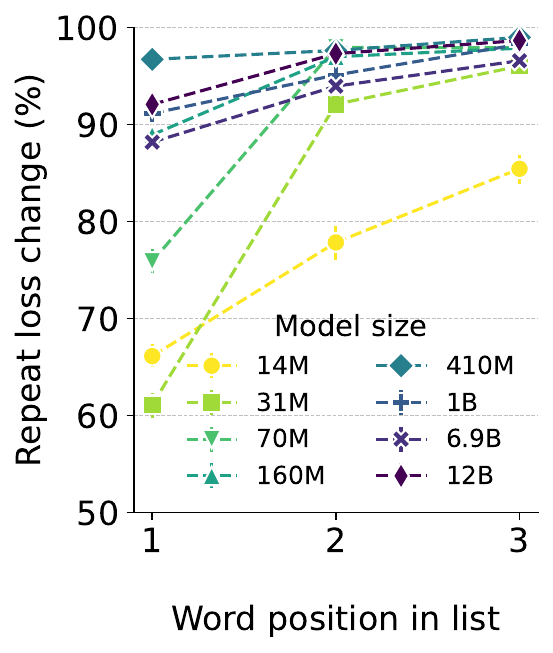}
        \caption{Retrieval improves for items later in the list (fully-trained models).}
        \label{fig:memory:per_token}
    \end{subfigure}
    \caption{\textbf{Retrieval of arbitrary nouns across time and scale.} Each data point represents the 20\% trimmed mean across $N=230$ observations, shaded areas/error bars are 95\% confidence intervals (bootstrap).}
    \label{fig:memory}
\end{figure*}

Evaluating an LM on the full retrieval evaluation suite yields one retrieval score (repeat loss change) per each input vignette. The final retrieval score, per each training step and per model size, was obtained by taking an average across all (in this case $N = 230$) scores. To minimize the potential influence of outliers in averaging, we used the 20\% trimmed mean \cite{wilcox_modern_2003} as the aggregating metric. The results of this experiment are reported in Figure \ref{fig:memory}.

\paragraph{Experiment 2: Correlations with zero-shot benchmark learning.} To test how learning of verbatim in-context retrieval relates to the learning of zero-shot benchmark tasks assessing text understanding, we collected the zero-shot evaluation results on various NLP benchmarks that were available for the Pythia suite of LMs\footnote{\url{https://github.com/EleutherAI/pythia/tree/main/evals/pythia-v1}}. Evaluations were available for the following 6 model sizes: \{160M, 410M, 1.4B, 2.8B, 6.9B, 12B\} and across 27 checkpoints\footnote{Checkpoints corresponding to the following Pythia training steps were evaluated: \{0, 1, 2, 4, 8, 16, 32, 64, 128, 256, 512, 1000, 3000, 13000, 23000, 33000, 43000, 53000, 63000, 73000, 83000, 93000, 103000, 113000, 123000, 133000, 143000\}.} during training, starting with the initial and ending with the fully-trained model. All individual tasks ($N = 65$) used accuracy as the final metric. The main groups of tasks used in the experiment are summarized in Table \ref{tab:benchmarks}. See Table \ref{tab:benchmarks_full}, Appendix \ref{sec:appendix} for the full task list.

For each benchmark task (e.g. Lambada, SciQ etc.), we computed the correlation $\rho^{traj} = \mathrm{Spearman}(S^{ret}, S^{bench})$ between the learning trajectory of the benchmark task $S^{bench}$ (i.e. task performance scores across the 27 checkpoints) and the learning trajectory of our verbatim retrieval effect $S^{ret}$ (i.e. repeat loss change $L^r$ across the same 27 checkpoints). The Massive multitask understanding benchmark \citep[MMLU,][]{hendrycks_measuring_2021} consists of an array of domain-specific exams (e.g. marketing, clinical knowledge, nursing) which are grouped into 4 higher-level categories (humanities, STEM, social sciences, and `other (business, health, misc.)', see Table \ref{tab:benchmarks_full}, Appendix \ref{sec:appendix}). For these grouped tasks, we first averaged the learning trajectories per each group and then correlated them with verbatim retrieval effect.  We used the rank-based Spearman correlation coefficient where a value of 1 indicates a perfect monotonically increasing relationship between two variables and is robust to any deviations from normality in data distributions.

\paragraph{Experiment 3: Effect of noun concreteness on retrieval.} To test for retrieval of concrete and abstract nouns, we evaluated LMs on the same paradigm as in the first experiment, but the noun lists were composed of either concrete or abstract nouns. We used abstract and concrete English nouns collected by \citet{brysbaert_concreteness_2014} where human participants were asked to indicate ``\textit{how concrete the meaning of each word is for you}'' by rating each noun on a 5-point rating scale ranging from 1 ``abstract (language-based)'' to 5 ``concrete (experience-based)''. Each word was rated by at least 25 participants and an average score across participants represents each noun's final rating.
\small
\input{concreteness_extremes_table}
\normalsize
\paragraph{Concreteness extremes} In our experiments, we used the ``concreteness extremes'' subset of the noun pool by \citet{schulte_im_walde_distributional_2022}. This subset contained the 500 nouns ranked as most concrete and 500 nouns ranked as most abstract. To give an idea, the topmost, mid and lowest ranked nouns for each category are shown in Table \ref{tab:abst-conc}. As in Experiment 1, each noun was presented in all ordinal positions to rule out any position-specific effects. Our final stimulus set contained, for each semantic category, $N = 498$ input sequences with lists of 3 nouns. 

\section{Results}

\begin{figure*}[!ht]
\includegraphics[width=1\textwidth]{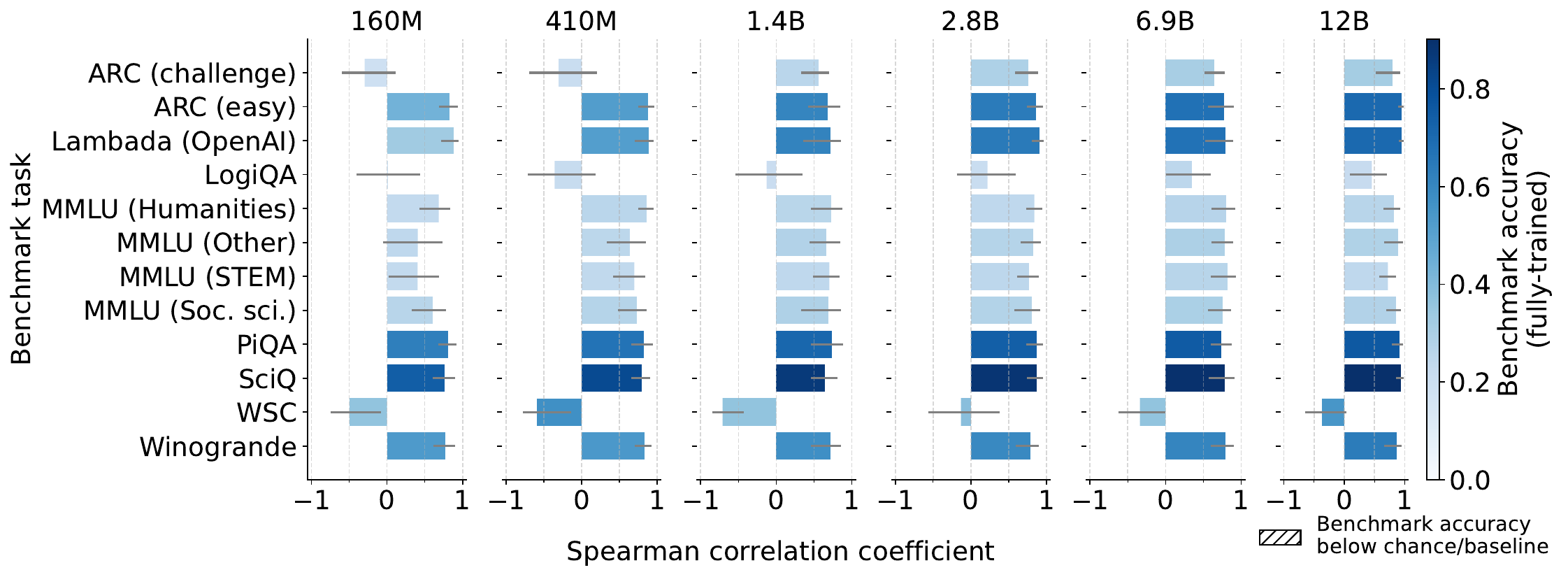}
\caption{\textbf{Correlations between the learning trajectories of verbatim retrieval and select benchmark tasks.} Error bars denote the 95\% confidence interval (bootstrap, $N = 5000$) around the correlation coefficient. Color-coded is the benchmark performance accuracy at the end of training for each task. Chance performance for ARC, LogiQA, MMLU*, LogiQA, SciQ is 0.25, for PiQA, WSC, and Winogrande 0.5. For LAMBADA we threshold against predicting a random in-context token (0.016) (\citealp[see Table 1 in][]{paperno_lambada_2016}). See Table \ref{tab:benchmarks} for task descriptions.}
\label{fig:benchmark_corr}
\end{figure*}

\begin{figure}
\includegraphics[width=1\linewidth]{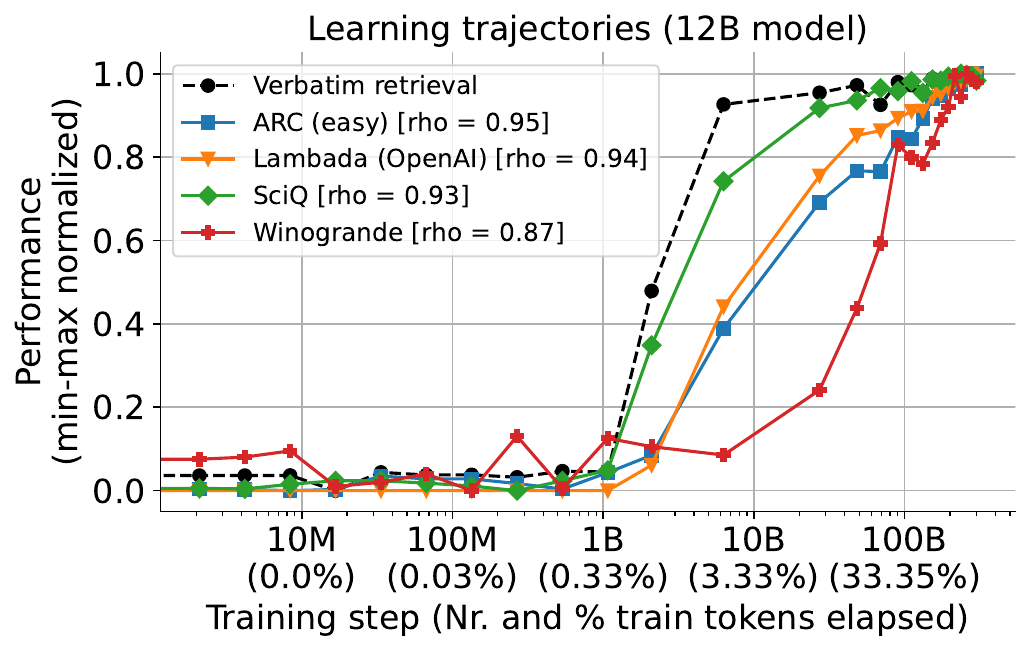}
\caption{Examples of learning trajectories (12B model) for tasks that showed strongest correlations with verbatim retrieval for the largest tested model. For visualization purposes, accuracy scores are min-max normalized to fall in the $[0, 1]$ range.}
\label{fig:learning_traj}
\end{figure}

\subsection{Verbatim retrieval across time and scale.} 

\paragraph{Verbatim retrieval learned early in training across model sizes.} All tested models, from the smallest (14 million parameters) to the largest (12 billion parameters), learned to retrieve verbatim repeated nouns (Fig. \ref{fig:memory:dynamics}). At the end of training, all models above 31 million parameters showed a near 100\% repeat loss change, indicating exact retrieval. The smallest two models (14M and 31M parameters) showed weaker, yet still substantial retrieval effect (around 80\% change in repeat loss).

Inspecting the dynamics of repeat loss change across training, we see that generally models learned verbatim retrieval early. After about 1B tokens (0.3\% of total dataset), the change in repeat loss starts increasing and, for all larger models, plateaus at approximately 4B tokens (less than 5\% of the total tokens in the dataset). The smallest two models had a slower learning curve as evidenced in the fact that their repeat loss change plateaued later, after roughly 20B tokens.

To confirm that reduction in repeat loss was due to retrieval of the \textit{original nouns} and not due to LMs simply having more context when encountering nouns at the end of sequence or due to memorization of lists from training data, we evaluated the loss change in the same paradigm but where the nouns in the second list were unrelated to the nouns in the original list (i.e. there were no matching in-context nouns to retrieve). Fig. \ref{app:fig_mem_control} in Appendix \ref{sec:appendix:control} confirms that no important loss change occurred in this condition (loss change overall remained < 10\%), replicating the GPT-2 results by \citet{armeni_characterizing_2022} and indicating that the change of loss was specific to \textit{verbatim retrieval} of tokens from context.

\paragraph{Retrieval improves for nouns deeper in the list.} In the previous result, we reported repeat loss change aggregated over all three nouns in the list. Yet, nouns deeper in the list have an advantage because at that point the LM has seen strong evidence of repetition. Does retrieval performance depend on the position in the list?

\begin{figure*}[!ht]
\centering
    \begin{subfigure}[t]{0.46\textwidth}
        \includegraphics[width=\textwidth]{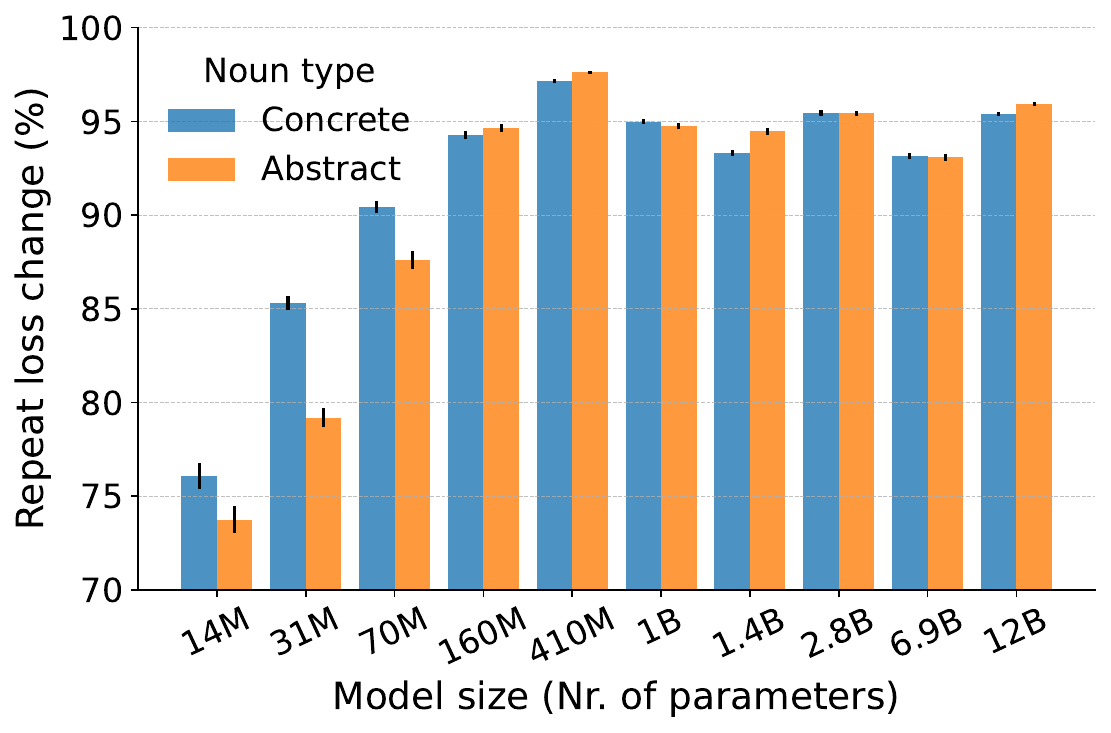}
        \caption{Concrete vs. abstract noun retrieval at the end of training. The three smallest models show a weak advantage for retrieving concrete nouns.}
        \label{fig:abst_conc:all}
    \end{subfigure}
    \hfill
    \begin{subfigure}[t]{0.52\textwidth}
        \includegraphics[width=\textwidth]{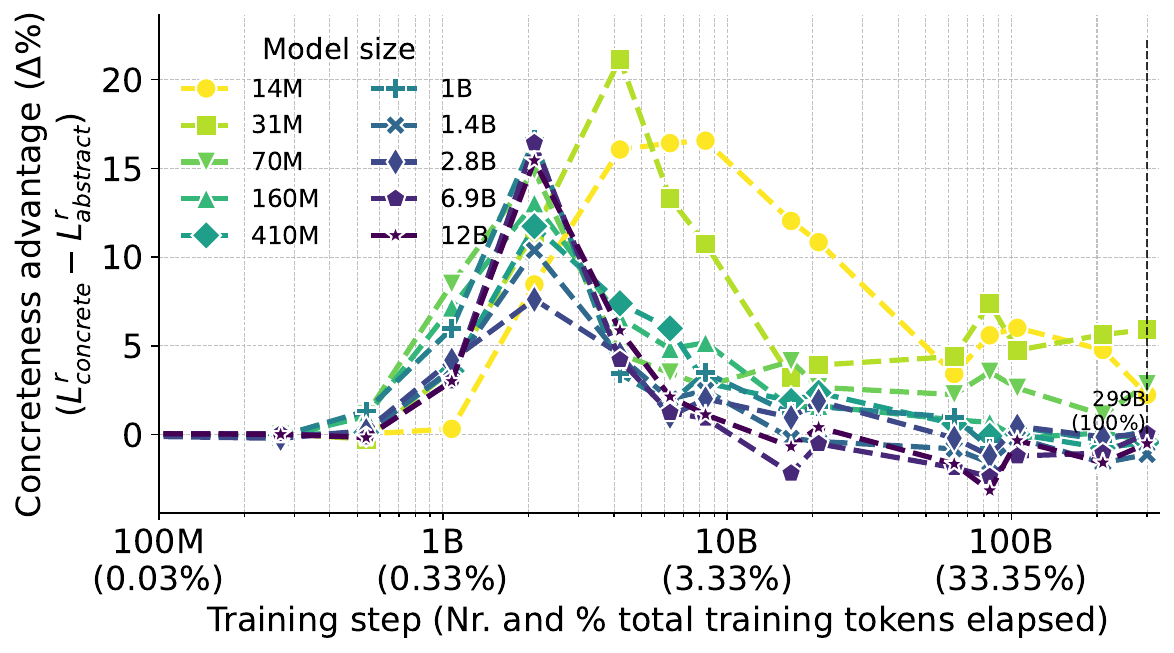}
        \caption{Retrieval of concrete vs. abstract nouns over the course of training. Around the transition point, all models show an advantage to retrieve concrete nouns.}
        \label{fig:abst_conc:dynamics}
    \end{subfigure}
    \caption{\textbf{Retrieval of concrete and abstract nouns}. \textbf{a)} Each bar shows the 20\% trimmed mean across $N=498$ observations, error bars show 95\% confidence intervals (bootstrap). See also Fig. \ref{fig:abst-conc-dist} in Appendix \ref{app:abst-conc-dist}. \textbf{b)} Each data point shows the difference in mean repeat loss change for concrete vs. abstract nouns (concreteness advantage).}
    \label{fig:abst_conc}
\end{figure*}

In Fig. \ref{fig:memory:per_token}, we report repeat loss change of fully-trained models broken down per noun position within the list. Retrieval indeed becomes better later in the list. While all models show this trend, the position-specific advantage is more pronounced for the smaller models (14M, 31M, and 70M). For example, the 70M model shows 62\% repeat loss change on the first and a 95\% change on the last token in the list. This indicates that subsequent repetitions reinforce the evidence that the model has entered a repeated list and is in line with recent results where next-word prediction performance of GPT-2 improved on spans of repeated text \citep{vaidya_humans_2023}.

\subsection{Correlations with benchmark task learning}

\paragraph{Learning of verbatim retrieval is positively correlated with zero-shot performance on more challenging benchmark tasks.} In Figure \ref{fig:benchmark_corr} we show the results of the correlation experiment. Generally, most tasks showed a positive correlation with the learning of verbatim retrieval. The correlations and their reliability, as well as the benchmark accuracy itself, tended to increase as the models grow in size, showing that the larger models were more robust learners overall. The highest correlations were observed for the Lambada, PiQA, SciQ, and ARC (easy) benchmarks. For example, the largest 12B model (Figure \ref{fig:learning_traj}) showed a near perfect rank correlation ($\rho \simeq 0.95$) on the four tasks. These are also the tasks where the model showed generally the highest performance accuracy at the end of training.

For the Winograd schema challenge, LogiQA, and the hard version of the AI2 reasoning challenge, the correlation estimates were generally unstable, likely because the performance on these benchmarks was lower to begin with. That is, even though all the models were able to retrieve verbatim in-context tokens, they failed to solve the respective benchmarks in zero-shot settings.

\subsection{Effect of noun concreteness on retrieval}

\paragraph{Concreteness retrieval advantage observed early during training.} Overall, all models learned to retrieve either abstract or concrete nouns. Repeat loss change at the end of training (on either repeated concrete or abstract nouns) was generally high and ranged from 73\% (14M model) to around 95\% for models larger than 160M parameters (Fig. \ref{fig:abst_conc:all}). The 14M, 31M and 70M models showed better retrieval for concrete nouns. The effects, although detectable, were small --- on average the relative loss change for concrete nouns is greater by between 2\% and 6\% compared to abstract nouns (see also Fig. \ref{fig:abst-conc-dist}, Appendix \ref{app:abst-conc-dist} for visualizations of full distributions).

To test whether nouns semantics affected retrieval during training, we computed the difference in average repeat loss change between concrete and abstract nouns $\Delta L^{r} = \bar{L}^{r}_{concrete} - \bar{L}^{r}_{abstract}$ across the training checkpoints. The difference curves in Fig. \ref{fig:abst_conc:dynamics} show that around the transition point (1-2B tokens intro training), when LMs begin to learn the retrieval, concrete nouns showed 7-17\% greater change in repeat loss meaning they were easier to retrieve than abstract nouns. The concreteness advantage occurred in all models and the smallest models (14 and 70M parameters) showed the largest effects.

\section{Discussion}

We showed that transformer LMs learned verbatim retrieval in a sudden transition, early in training, with the performance remaining stable over the course of training. The sharp onset of retrieval capacity around 1-2B tokens in training (approximately 1\% total training data) is in line with the results reported by \citet{olsson_-context_2022} who showed that the LM loss over in-context tokens started dropping suddenly 1-2\% tokens in training (between 2.5B and 5B tokens). Once the learning change had occurred, the LMs became better at predicting repeated text --- which is what was tested in the current work.

The learning trajectory of verbatim retrieval also coincides with the LMs' learning trajectories on zero-shot benchmarks. This was reflected in the generally high and robust correlations across training for select tasks in our results. Specifically, an abrupt change around 1B tokens in training was observed in the task of predicting the last token of a narrative passage (Lambada, \citealt{paperno_lambada_2016}), multiple choice exams (SciQ, \citealt{welbl_crowdsourcing_2017}, ARC Reasoning Challenge, \citealt{clark_think_2018}), and in the Winogrande benchmark \citep{sakaguchi_winogrande_2021} which requires pronoun resolution based on common-sense reasoning.

Retrieving in-context information (e.g. lists of nouns) verbatim is a basic computation needed for solving a zero-shot multiple-choice task: given a prompt with only in-context instructions (that is, the question and the list of possible answers), an LM system must index and retrieve (i.e. increase the probability of) the token representing the correct answer. In this sense, retrieving the correct in-context tokens is a necessary step. It is evident, however, that it is not sufficient and that verbatim retrieval must be learned along with other computations.

Consider the Lambada and Winogrande benchmarks, where the task is to predict the passage- or sentence-final word which itself is not predictable on the basis of immediately preceding words. To take an example from the Winogrande benchmark: ``\textit{Robert woke up at 9am while Samuel woke up at 6am, so \textbf{he} had less time to get ready for school}''\footnote{\url{https://winogrande.allenai.org/}}. The task is to answer who the pronoun ``he'' refers to (Robert or Samuel). To this end, the LM must first establish that 9am is later than 6am --- a distinct computational step indicating that ``he'' refers to ``Robert'' --- and only then retrieve the name to be predicted as the response.

In the final experiment, we show that around the transition point (after $\simeq$1B training tokens), when the capacity for verbatim retrieval occurs, noun semantics affect the retrieval --- models showed an advantage to retrieve concrete, as opposed to abstract nouns. Why would LM in-context retrieval be sensitive to noun semantics? 

In humans, concrete words, especially nouns, tend to be acquired earlier in development compared to abstract words \citep{gleitman_hard_2005}. This advantage is presumably conferred by hearing words for concrete objects and concurrently observing or interacting with the objects the heard words refer to in the world. LMs as text-based statistical learners by construction have no direct access to word semantics via experience or text-external data \citep{bisk_experience_2020}. Nevertheless, text statistics, governed by human language use, can serve as a cue to the semantic structure of language --- in this case, the lexicon. It is an empirical question whether and what aspects of the linguistic system are in fact recovered by LMs in the service of the next-word prediction objective and subsequently reflected in the LM behavior or internal mechanisms \cite{manning_human_2022, pavlick_symbols_2023}.

We speculate that earlier in training, LMs are leveraging the fact that concrete nouns tend to be used in more predictable, less diverse contexts \citep{schulte_im_walde_distributional_2022} where presumably token repetition would be more likely to occur. However, once the LMs and the training compute scale in size, this distributional difference no longer confers an important advantage for retrieval. The phenomenon of concreteness advantage early, but not later in training underscores the general notion that with the increasing amounts of training data, LMs as machine learning systems become incommensurate with human learners \citep[see also][]{vaidya_humans_2023}, who operate on the order(s) of magnitude smaller amount of learning data, at least in terms of number of words --- recent estimates point to around 100M words by adolescence \citep{warstadt_what_2022}.

\paragraph{Future work.} In this study we investigated retrieval across a diverse set of nouns, and broken down by a core semantic dimension. However, LMs are statistical learners. A dimension of future work will be to disentangle the \textit{learning sources} that LMs leverage to perform retrieval. In a recent study, \citet{yu_characterizing_2023} showed that pretraining frequency can override the retrieval of counterfactual in-context information. An LM is more likely to predict a proper noun that was frequently occurring in pretraining, e.g. ``\textit{Warszaw}'', even when the counterfactual in-context prompt suggests it should retrieve a different name ( ``\textit{The capital of Poland is London. What is the capital of Poland?}''). Our present results do not speak directly to this issue as our paradigm does not involve counterfactuals. It is based on lists of arbitrary nouns that unlikely frequently co-occurred in pretraining data. However, it would be important to establish whether and to what extent the in-context retrieval in general is governed by the pretraining frequencies of individual common nouns and to what degree this capacity is robust to pretraining statistics.

Finally, our measure of verbatim retrieval is a behavioral measure insofar that it only takes into account the output of the LM. The field of model interpretability has seen an increased interest in recent years and aims to reverse engineer the computations of LMs \citep[e.g.][among others]{olsson_-context_2022, elhage_mathematical_2021, wang_interpretability_2023, zhang_towards_2023}. Future work could focus on investigating the internal mechanisms and their causal role in transformer in-context retrieval. There is consistent evidence suggesting that LMs develop dedicated attention heads \citep{olsson_-context_2022, wang_interpretability_2023, yu_characterizing_2023, vaidya_humans_2023} governing the retrieval capacity. Whereas this line work frequently focuses on interpretability for practical purposes (e.g. better control of LM output in downstream applications), it would be valuable to simultaneously develop a more fine-grained computational characterization of LM mechanisms interpretable with respect to cognitive science constructs like the short-term memory \cite{cowan_many_2017}.

In cognitive neuroscience, language features derived from transformer LMs (contextualized word embeddings) are currently among the best performing when it comes to predicting brain data recorded in human language processing tasks \citep[e.g.][]{schrimpf_neural_2021, goldstein_shared_2022, caucheteux_brains_2022}. However, these high-dimensional features and the resulting statistical fits are frequently hard to interpret. Coupled with loose theoretical motivations such high predicting models can be right for the wrong scientific reasons \citep[see e.g.][]{antonello_predictive_2023}. A better characterization of LM mechanisms in terms of cognitive capacities \citep[e.g.][]{lakretz_can_2022} would be instrumental in understanding how and why LMs succeed in modeling human brain and cognitive data.

\section{Conclusion}

Retrieving information from context is an important capacity of transformer language models. In this work, we investigated how the ability to retrieve repeated nouns from context develops across LM training and scale and its dependence on whether the retrieved nouns denote concrete or abstract entities. Retrieval was learned early in training across scale and once learned, it remained stable. Retrieval learning was robustly correlated with learning of zero-shot task performance. Around the point when the in-context retrieval was learned, the models showed advantage to retrieving concrete as opposed to abstract nouns and the advantage dissipated as the models saw more training data.

\section{Limitations}

There are certain limitations to current work. While our test suite was designed to test arbitrary target nouns, we did not investigate whether and how well LM retrieval generalizes to other parts of speech (say, to verbs, adjectives). Similarly, the currently reported paradigm relies on a single vignette, it would be important to use a more diverse set of vignettes to confirm that the results generalize across topic domains. However, given the robustness and size of the effect here and in past reports by others, it is likely that the finding would generalize across a diversity of vignettes. Finally, our results are limited to English, which currently the dominant language in terms of available resources in language technologies. Extending the study to other languages with, for example, different grammatical properties (e.g. richer noun morphology) or less resources would be a welcome effort.

\section*{Data and Code Availability}

The code used to run the experiments is available at: \url{https://github.com/KristijanArmeni/verbatim-memory-in-NLMs}

\noindent
The materials and data used in the experiments are available at: \url{https://doi.org/10.17605/OSF.IO/A6GSW}

\section*{Computational requirements}

Experiments described in this report were run on the A100 Nvidia GPU nodes on an high-performance computing cluster (HPC). To evaluate the smallest (14M) model, we requested 8GB of RAM and the evaluation completed on the order of a few minutes. RAM requirements were progressively increased to evaluate larger models. For the largest (12B) model, we requested 80GB of RAM and the evaluation completed in about 30 minutes.

\section*{Acknowledgements}

KA would like to thank Christopher Honey and Tal Linzen for scientific and administrative support. This work was supported in part through the NYU IT High Performance Computing resources, services, and staff expertise. KA, MP, and SP want to thank Rebecca Swisdak, Mili Bauer and Živa Antauer for exceptional administrative support. 
This work was supported by the Slovenian Research and Innovation Agency via the bilateral research project Working Memory based assessment of Large Language Models (BI-US/22-24-170), the research project Embeddings-based techniques for Media Monitoring Applications (L2-50070, co-funded by the Kliping d.o.o. agency) and the core research programme Knowledge Technologies (P2-0103).

\bibliography{library.bib}

\appendix

\section{Appendix A}
\label{sec:appendix}

\begin{figure*}[!ht]
\centering
\includegraphics[width=0.7\textwidth]{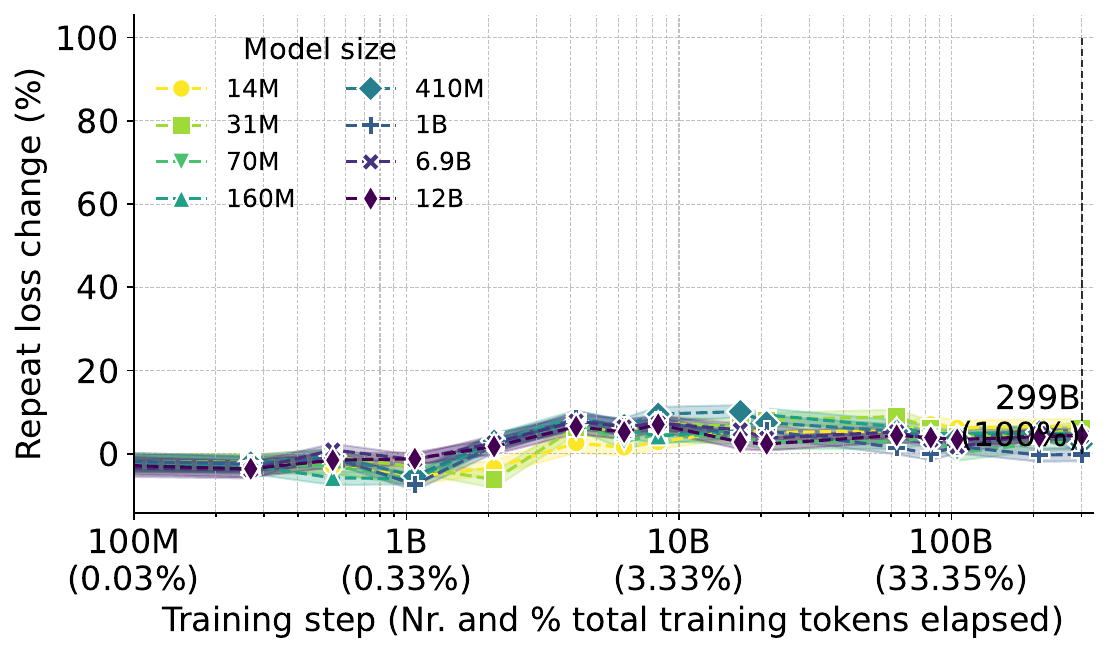}
\caption{Evaluating repeat loss change in a control condition where there were no verbatim repeated in-context nouns (hence, no retrieval was possible). Each data point shows the 20\% trimmed mean across $N=230$ observations, shaded areas/error bars are 95\% confidence intervals (bootstrap).}
\label{app:fig_mem_control}
\end{figure*}

\begin{figure*}[!ht]
\centering
\includegraphics[width=\linewidth]{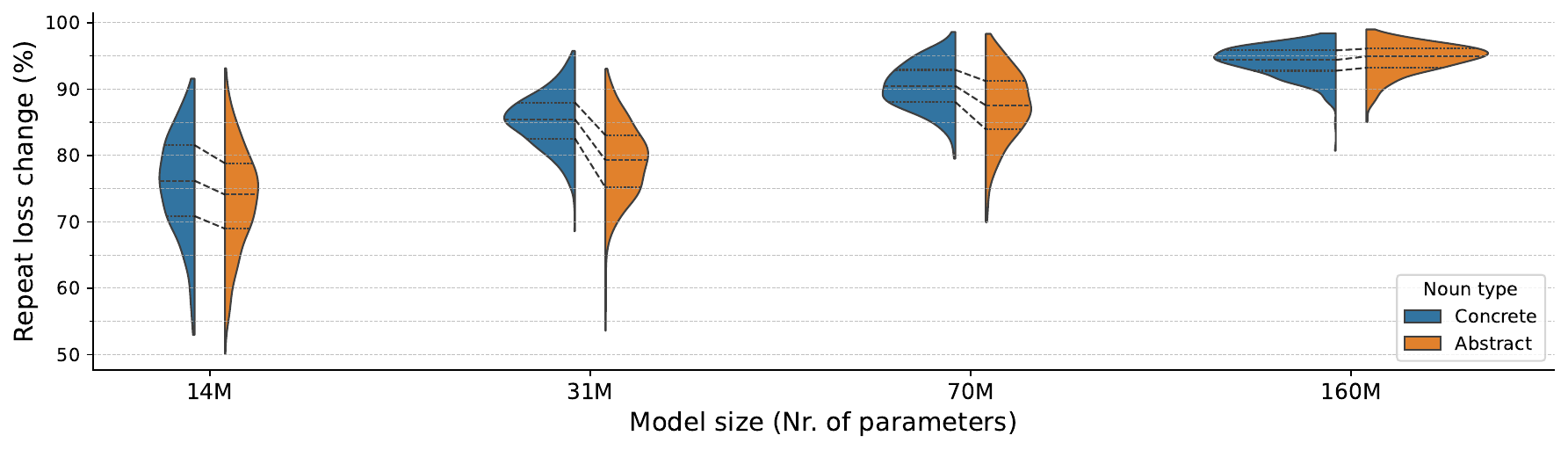}
\caption{Data distributions comparing retrieval scores for concrete and abstract nouns for 4 smallest models from Fig. \ref{fig:abst_conc}. Each violin plot KDE density estimates over $N = 498$ data points. The inner lines show the first, second (median) and the third quartiles of the distribution.}
\label{fig:abst-conc-dist}
\end{figure*}

\subsection{Retrieval control} \label{sec:appendix:control}

Figure \ref{app:fig_mem_control} shows memory retrieval results when nouns are not repeated.

\subsection{Abstract vs. concrete data distributions} \label{app:abst-conc-dist}

The violin plots in Fig. \ref{app:abst-conc-dist} show the distributions underlying respective bar plots in Fig. \ref{fig:abst_conc:all}.

\subsection{Zero-shot benchmark tasks overview} \label{sec:appendix:zero-shot}

The full list of benchmark tasks used in Experiment 2 is provided in Table \ref{tab:benchmarks_full}.

\small
\input{pythia_benchmarks}

\end{document}

%% file: concreteness_extremes_table.tex
\begin{table}[!h]
\small
\caption{The topmost, mid and lowest ranked words and their concreteness ratings for the concrete and abstract noun pool.}
\label{tab:abst-conc}
\begin{tabular}{llrlr}
\toprule
 & \multicolumn{2}{c}{\textbf{Concrete}} & \multicolumn{2}{c}{\textbf{Abstract}} \\
 Rank & Word & Rating & Word & Rating \\
\midrule
1 & whisky & 5.00 & oneness & 1.96 \\
250 & canister & 4.93 & respite & 1.77 \\
500 & eyebrow & 4.85 & spirituality & 1.07 \\
\bottomrule
\end{tabular}
\end{table}

%% file: pythia_benchmarks.tex
\begin{table*}
\small
\caption{Benchmark categories for the Pythia models. The Task Key column corresponds to the task key used in the Pythia evaluation files (\url{https://github.com/EleutherAI/pythia/tree/main/evals/pythia-v1}).}
\begin{tabular}{llll}
\toprule
 & Benchmark Name & Benchmark Subcategory & Task Key \\
\midrule
1 & ARC (challenge) & None & arc\_challenge \\
2 & ARC (easy) & None & arc\_easy \\
3 & Lambada (OpenAI) & None & lambada\_openai \\
4 & LogiQA & None & logiqa \\
5 & MMLU & MMLU (Soc. sci.) & mmlu\_high\_school\_government\_and\_politics \\
6 & MMLU & MMLU (Soc. sci.) & mmlu\_sociology \\
7 & MMLU & MMLU (Other) & mmlu\_business\_ethics \\
8 & MMLU & MMLU (Other) & mmlu\_medical\_genetics \\
9 & MMLU & MMLU (STEM) & mmlu\_high\_school\_physics \\
10 & MMLU & MMLU (Other) & mmlu\_professional\_medicine \\
11 & MMLU & MMLU (Other) & mmlu\_miscellaneous \\
12 & MMLU & MMLU (STEM) & mmlu\_college\_physics \\
13 & MMLU & MMLU (Humanities) & mmlu\_professional\_law \\
14 & MMLU & MMLU (Humanities) & mmlu\_high\_school\_world\_history \\
15 & MMLU & MMLU (Other) & mmlu\_global\_facts \\
16 & MMLU & MMLU (Humanities) & mmlu\_high\_school\_us\_history \\
17 & MMLU & MMLU (Other) & mmlu\_marketing \\
18 & MMLU & MMLU (Soc. sci.) & mmlu\_high\_school\_microeconomics \\
19 & MMLU & MMLU (Other) & mmlu\_college\_medicine \\
20 & MMLU & MMLU (Soc. sci.) & mmlu\_human\_sexuality \\
21 & MMLU & MMLU (STEM) & mmlu\_electrical\_engineering \\
22 & MMLU & MMLU (STEM) & mmlu\_elementary\_mathematics \\
23 & MMLU & MMLU (STEM) & mmlu\_high\_school\_chemistry \\
24 & MMLU & MMLU (Other) & mmlu\_professional\_accounting \\
25 & MMLU & MMLU (Humanities) & mmlu\_world\_religions \\
26 & MMLU & MMLU (STEM) & mmlu\_machine\_learning \\
27 & MMLU & MMLU (Soc. sci.) & mmlu\_high\_school\_psychology \\
28 & MMLU & MMLU (Humanities) & mmlu\_moral\_scenarios \\
29 & MMLU & MMLU (STEM) & mmlu\_high\_school\_computer\_science \\
30 & MMLU & MMLU (Soc. sci.) & mmlu\_security\_studies \\
31 & MMLU & MMLU (STEM) & mmlu\_computer\_security \\
32 & MMLU & MMLU (Humanities) & mmlu\_high\_school\_european\_history \\
33 & MMLU & MMLU (STEM) & mmlu\_college\_computer\_science \\
34 & MMLU & MMLU (Soc. sci.) & mmlu\_econometrics \\
35 & MMLU & MMLU (STEM) & mmlu\_college\_mathematics \\
36 & MMLU & MMLU (Other) & mmlu\_clinical\_knowledge \\
37 & MMLU & MMLU (Soc. sci.) & mmlu\_professional\_psychology \\
38 & MMLU & MMLU (Other) & mmlu\_nutrition \\
39 & MMLU & MMLU (STEM) & mmlu\_abstract\_algebra \\
40 & MMLU & MMLU (Humanities) & mmlu\_logical\_fallacies \\
41 & MMLU & MMLU (STEM) & mmlu\_astronomy \\
42 & MMLU & MMLU (STEM) & mmlu\_high\_school\_mathematics \\
43 & MMLU & MMLU (STEM) & mmlu\_high\_school\_biology \\
44 & MMLU & MMLU (Soc. sci.) & mmlu\_high\_school\_geography \\
45 & MMLU & MMLU (Other) & mmlu\_anatomy \\
46 & MMLU & MMLU (Humanities) & mmlu\_jurisprudence \\
47 & MMLU & MMLU (Other) & mmlu\_management \\
48 & MMLU & MMLU (Humanities) & mmlu\_prehistory \\
49 & MMLU & MMLU (STEM) & mmlu\_college\_biology \\
50 & MMLU & MMLU (Humanities) & mmlu\_moral\_disputes \\
51 & MMLU & MMLU (STEM) & mmlu\_high\_school\_statistics \\
52 & MMLU & MMLU (Soc. sci.) & mmlu\_us\_foreign\_policy \\
53 & MMLU & MMLU (Other) & mmlu\_human\_aging \\
54 & MMLU & MMLU (STEM) & mmlu\_college\_chemistry \\
55 & MMLU & MMLU (Other) & mmlu\_virology \\
56 & MMLU & MMLU (Soc. sci.) & mmlu\_public\_relations \\
57 & MMLU & MMLU (STEM) & mmlu\_conceptual\_physics \\
58 & MMLU & MMLU (Soc. sci.) & mmlu\_high\_school\_macroeconomics \\
59 & MMLU & MMLU (Humanities) & mmlu\_international\_law \\
60 & MMLU & MMLU (Humanities) & mmlu\_philosophy \\
61 & MMLU & MMLU (Humanities) & mmlu\_formal\_logic \\
62 & PiQA & None & piqa \\
63 & SciQ & None & sciq \\
64 & Winogrande & None & winogrande \\
65 & WSC & None & wsc \\
\bottomrule
\label{tab:benchmarks_full}
\end{tabular}
\end{table*}